\documentclass{article}

    \PassOptionsToPackage{numbers, compress}{natbib}

    \usepackage[preprint]{neurips_2023}

\usepackage[colorlinks,linkcolor=blue]{hyperref}
\usepackage[dvipsnames]{xcolor}
\definecolor{cite_color}{RGB}{190,0,60}
\definecolor{link_color}{RGB}{0,102,102}
\definecolor{link_color}{RGB}{153, 0,0}  
\definecolor{url_color}{RGB}{153, 102,  0}
\definecolor{emp_color}{RGB}{0,0,255}
\hypersetup{
 colorlinks,
 citecolor=cite_color,
 linkcolor=link_color,
 urlcolor=url_color}

\usepackage[utf8]{inputenc} 
\usepackage[T1]{fontenc}    
\usepackage{url}            
\usepackage{booktabs}       
\usepackage{amsfonts}       
\usepackage{nicefrac}       
\usepackage{microtype}      
\usepackage{makecell}
\usepackage{wrapfig}
\usepackage{caption}

\usepackage{microtype}
\usepackage{graphicx}
\usepackage{subfigure}
\usepackage{booktabs} 

\usepackage{amsmath}
\usepackage{amssymb}
\usepackage{mathtools}
\usepackage{amsthm}
\usepackage[numbers]{natbib}

\usepackage[capitalize,noabbrev]{cleveref}

\usepackage{color}
\usepackage{colortbl}
\usepackage{diagbox}

\usepackage{etoc}
\etocdepthtag.toc{mtchapter}
\etocsettagdepth{mtchapter}{subsection}
\etocsettagdepth{mtappendix}{none}

\usepackage{textcomp}
\usepackage{framed}
\usepackage{color}
\definecolor{shadecolor}{rgb}{0.94, 0.97, 1.0}

\usepackage{makecell}
\usepackage{enumitem,multirow}

\usepackage{harpoon}

\theoremstyle{plain}

\theoremstyle{definition}

\theoremstyle{remark}

\title{Molecule Generation for Drug Design: \\a Graph Learning Perspective}

\begin{document}

\author{%
  Nianzu Yang$^\dag$, Huaijin Wu$^\dag$, Kaipeng Zeng, Yang Li, Junchi Yan\thanks{Junchi Yan is the correspondence author.} \\
  Department of Computer Science and Engineering\\
  MoE Key Lab of Artificial Intelligence\\
  Shanghai Jiao Tong University\\
  \texttt{\{yangnianzu,whj1201,zengkaipeng,yanglily,yanjunchi\}@sjtu.edu.cn} \\
}

\renewcommand{\thefootnote}{\fnsymbol{footnote}}
\footnotetext[2]{The first two authors contribute equally to this paper.}
\renewcommand{\thefootnote}{\arabic{footnote}}

\maketitle

\begin{abstract}
Machine learning, particularly graph learning, is gaining increasing recognition for its transformative impact across various fields. One such promising application is in the realm of molecule design and discovery, notably within the pharmaceutical industry. Our survey offers a comprehensive overview of state-of-the-art methods in molecule design, particularly focusing on \emph{de novo} drug design, which incorporates (deep) graph learning techniques. We categorize these methods into three distinct groups: \emph{i)} \emph{all-at-once}, \emph{ii)} \emph{fragment-based}, and \emph{iii)} \emph{node-by-node}. Additionally, we introduce some key public datasets and outline the commonly used evaluation metrics for both the generation and optimization of molecules. In the end, we discuss the existing challenges in this field and suggest potential directions for future research.
\end{abstract}

\section{Introduction}

In recent years, machine learning based drug discovery has been more and more conspicuous since it greatly reduces  time, money and labor costs for developing novel drugs~\citep{jimenez2020drug,zhu2020big,kim2020artificial}. Among the processes of drug development, generating chemical molecules with good quality and optimizing chemical molecules for desired properties are of particular importance. So the challenge lies in how to apply machine learning methods to generate ``good" molecules with or without additional constraints. Different approaches and models have been designed till now, including including those based on variational autoencoder (VAE)~\citep{kingma2014auto}, generative adversarial networks (GAN)~\citep{goodfellow2014generative}, reinforcement learning (RL)~\citep{kaelbling1996reinforcement}, and more.

To characterize molecules, several types of molecular representations are devised, ranging from 
Simplified Molecular Input Line Entry System (SMILES) strings~\citep{weininger1988smiles}
to manually predefined molecular features~\citep{redkar2020machine}. Among them, SMILES-based and graph-based representation methods are the most widely used in molecular generation tasks.
Early molecular generation methods are SMILES-based. SMILES can be seen as a type of 1D text representation. These SMILES-based methods cannot ensure 100\% chemical validity~\citep{elton2019deep} unless complicated constraints are added.
Meanwhile, molecules can naturally be represented using graphs, which are essentially a type of 2D representation. Recently, an increasing number of methods have shifted towards graph-based approaches. Unlike the 1D SMILES-based methods, 2D molecule generation approaches can easily ensure that the generated molecules are 100\% chemically valid. Additionally, graph-based representation has the ability to accurately depict the inherent structure of molecules. In light of the fact that graph-based representation is currently the mainstream method, this study will specifically concentrate on existing graph-based methodologies.

There also exist surveys on molecule generation and optimization.~\citep{guo2022systematic} and~\citep{faez2021deep} both provide a comprehensive overview of the literature in the field of deep generative models for graph generation. But they do not focus on molecules only. Apart from molecules, they also present deep generative models designed for other domains, such as social networks. 
As for~\citep{elton2019deep,alshehri2020deep}, they both put emphasis on the four architectures often utilized for molecule design methods. It is also worth mentioning
that~\citep{elton2019deep,alshehri2020deep,guo2022systematic,faez2021deep} all discuss molecule design methods based on different molecular representations, including SMILES, 2D representation referring to connectivity graph and 3D representation that contains coordinates of the atoms within a molecule. 
As previously noted, despite the extensive literature on SMILES-based generative models~\citep{gomez2018automatic,kang2018conditional,putin2018adversarial,grisoni2020bidirectional,griffiths2020constrained}, these methods have fallen out of mainstream use. Additionally, there is a noticeable disparity in the volume of research between 3D-generative methods for molecules and the more prevalent 2D graph-based methods. For readers interested in 3D-based methods, existing representative works like G-SchNet~\citep{gebauer2019symmetry}, E-NF~\citep{garcia2021n}, GEN3D~\citep{roney2022generating} and G-SphereNet~\citep{luo2022an} are recommended.

Different from the existing surveys mentioned above, our survey solely focuses on the drug design tasks and conducts a comprehensive overview of the state-of-the-art molecular design methods based on the 2D representation, i.e., only from a graph learning perspective. Additionally, compared to other surveys and an earlier version of our own survey, we include some more recent methods, particularly some diffusion-based methods, which were not covered in previous surveys.

In this survey, we provide a comprehensive review of the latest graph-based methods for molecule generation and optimization. These methods are classified into three categories based on their generation strategies: \emph{all-at-once}, \emph{fragment-based}, and \emph{node-by-node}. The survey also covers key public datasets and standard evaluation metrics used in this field. Additionally, we delve into an in-depth analysis of the current challenges faced in this area and proposes three promising directions for future research.

\section{Preliminaries and Problem Formulation}
\label{sec:form}
In this section, we first introduce graph-based molecule representation. Then, we formally formulate the problem of molecule generation task.

\subsection{\textbf{Graph-based Molecule Representation}}

In the field of Graph-based molecule representation~\citep{yang2022learning,guo2023survey,yang2023molerec}, it is common to use
a graph $G = (V, E)$ to model a molecule, where $V$ is the graph's node set mapping to atoms constituting a molecule and $E$ is denotes the graph's edge set mapping to chemical bonds, with $|V| = n$ and $|E| = m$. In molecule graph, nodes are sometimes representing atomic types from the periodic table, or representing certain kinds of molecule fragments. The node feature matrix $\mathbf{X}$ characterizes the property of each node while the adjacency matrix $\mathbf{A}$ characterizes the relationships between each node. Let the number of edge types be $b$ and the number of node types be $c$, then we have $\mathbf{A} \in \{0,1\}^{n\times n\times b}$ and $\mathbf{X} \in \{0, 1\}^{n \times c}$, where $\mathbf{A}_{ijk} = 1$ when there exists an edge with type $k$ between the $i^{th}$ and $j^{th}$ nodes, otherwise $0$. We can also represent the molecular graph using the node feature matrix $\mathbf{X}$ and the adjacency matrix $\mathbf{A}$, i.e., $G=(\mathbf{A},\mathbf{X})$.

\subsection{\textbf{Problem Formulation}}

Generation tasks aim to generate novel samples from a similar distribution as the training data~\citep{faez2021deep}.
A molecule generation method intends to generate novel, diverse molecules which follow the unknown data distribution $p(G)$ provided by a set of graphs $D_{G}$. A machine learning method towards this problem usually proposes a model to learn form large scales of data which either obtains an implicit strategy or estimates the $p(G)$ directly
and then samples from the distribution to generate new molecules.

\begin{figure}[tb!]
\centering
\includegraphics[width=\linewidth]{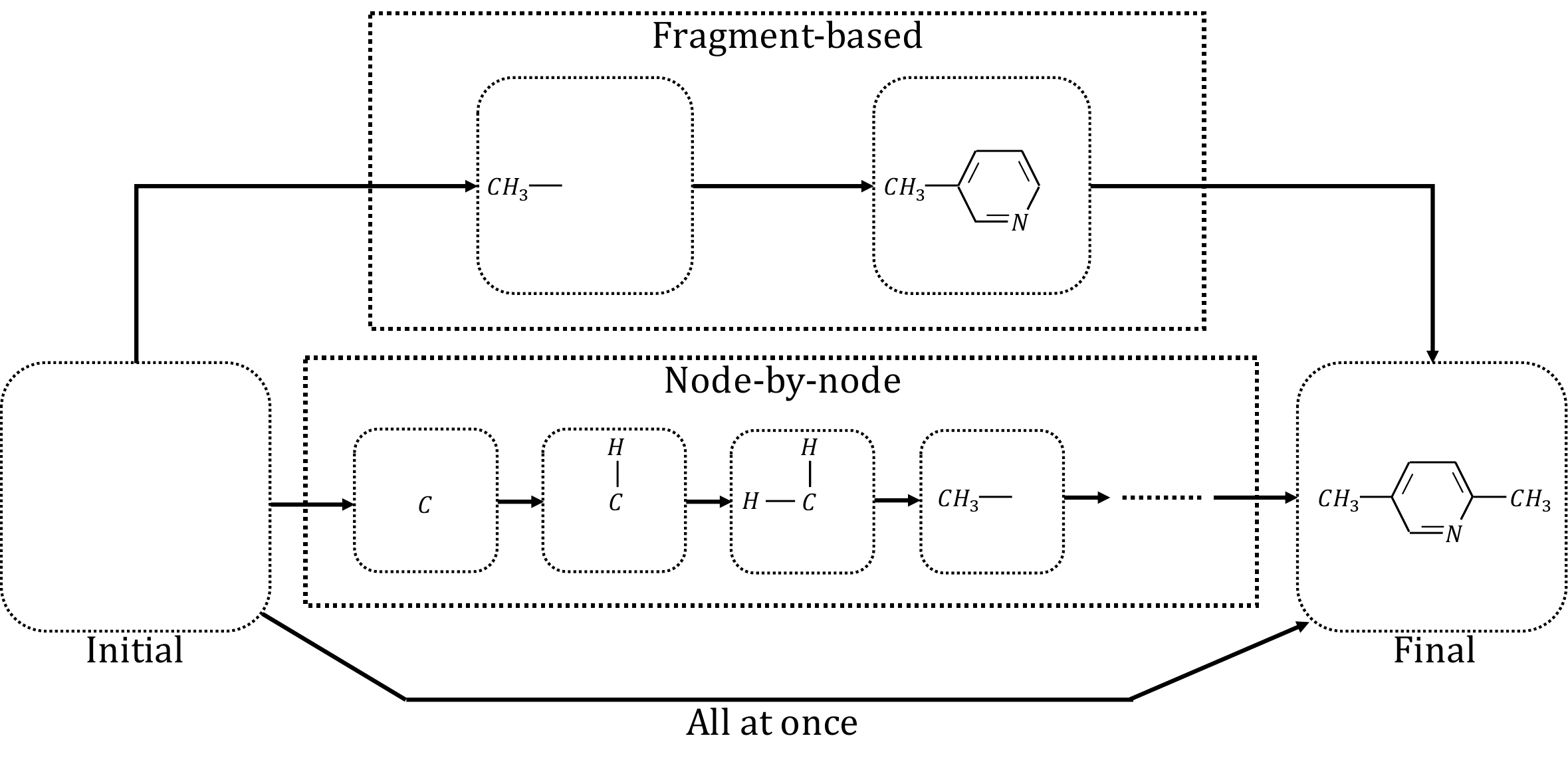}
\caption{Three typical molecule generation strategies.}
\label{fig:strategy}
\end{figure}

\section{Generation Strategies}
\label{sec:cate}
For classifying existing methods of \emph{de novo} molecule generation or molecule optimization, in this paper we propose to base on their generation granularity level as shown in Fig.~\ref{fig:strategy}: \emph{i)} using an all-at-one generation or optimization strategy; \emph{ii)} adopting rational substructures as editing blocks; or \emph{iii)} building graphs in a node-by-node setting. We describe their main features as summarized in Table~\ref{tab:models}.

\subsection{\textbf{Generation Strategy I: All-at-once}}
There are a number of deep graph generators to generate the entire molecular in one shot, which we call ``\emph{all-at-once}".

VGAE~\citep{kipf2016variational}, built upon a variational autoencoder (VAE)~\citep{kingma2014auto}, is a framework designed for unsupervised learning with graph-based data. VGAE leverages latent variables and is trained to learn interpretable latent representations, which are then used to generate new molecular graphs. Unlike VGAE, which can only learn from a single input graph, GraphVAE~\citep{simonovsky2018graphvae} 
is another VAE-based generative model that can learn from a set of graphs. The encoder of GraphVAE uses a graph convolutional network (GCN)~\citep{zhang2019graph} to embed the input molecular graph into a continuous representation $\mathbf{z}$ while the decoder of GraphVAE outputs a probabilistic fully-connected graph constrained by a predefined maximum size, from which discrete samples are drawn. However, GraphVAE encounters challenges in effectively aligning with the training data distribution and relies on an expensive graph matching procedure. In order to address these issues, MPGVAE~\citep{flam2020graph} integrates a message passing neural network (MPNN)~\citep{gilmer2017neural} to the encoder and decoder of the GraphVAE. Furthermore, \citep{ma2018constrained} proposes a specialized regularization framework for training VAEs that encourages the satisfaction of validity constraints for molecules, i.e. the number of bonding-electron pairs must not exceed the valence of an atom.

In addition to the aforementioned VAE-based methods, there are also approaches that utilize other generative models, such as MolGAN~\citep{de2018molgan} and GraphNVP~\citep{madhawa2019graphnvp}. MolGAN proposes an implicit generative model for molecular graphs, which adapts generative adversarial networks (GAN)~\citep{goodfellow2014generative} for graph-structured data and integrates a reinforcement learning objective to encourage the generation of molecules with desired properties.
GraphNVP is the first known molecule generation model based on invertible normalizing flow~\citep{dinh2017density,kingma2018glow}. It performs \emph{dequantization} technique~\citep{dinh2017density,kingma2018glow} to transform discrete adjacency tensor $\mathbf{A}$ and node label matrix $\mathbf{X}$ into continuous variables and then uses coupling layers to obtain latent representations $\mathbf{z}_\mathbf{A}, \mathbf{z}_\mathbf{X}$. 
$\mathbf{z}_\mathbf{A}$ and $\mathbf{z}_\mathbf{X}$ are concatenated together to obtain the final latent representation $\mathbf{z}$ of the molecule, i.e., $\mathbf{z} = \mathrm{CONCAT}(\mathbf{z}_\mathbf{A}, \mathbf{z}_\mathbf{X})$. After sampling a latent vector $\mathbf{z}$ from a known prior distribution and splitting $\mathbf{z}$ into $\mathbf{z}_\mathbf{A}$ and $\mathbf{z}_\mathbf{X}$, GraphNVP takes two steps to generate a molecule. The first is generating a graph structure $\mathbf{A}$ from $\mathbf{z}_\mathbf{A}$ and the second is generating node attributes $\mathbf{X}$ based on the structure $\mathbf{A}$ from $\mathbf{z}_\mathbf{X}$. Furthermore, GraphNVP trains a linear regressor on the latent space of molecules to quantitatively estimate a target chemical property. This enables the interpolation of the latent vector of a randomly selected molecule along the direction of the desired property, as learned through linear regression, thereby facilitating molecular optimization.

In recent developments, innovative approaches to molecular generation have emerged, building upon a class of advanced generative model known as diffusion models~\citep{ho2020denoising,song2021scorebased}. GDSS~\citep{jo2022score}, for instance, introduces a novel graph diffusion process to model the joint distribution of nodes and edges, utilizing a system of stochastic differential equations (SDEs). Subsequently, GDSS formulates specialized score matching objectives tailored to this diffusion process to estimate the gradient of the joint log density for each component. Moreover, it introduces a novel solver for the SDE system to facilitate efficient sampling from the reverse diffusion process. It is essential to emphasize that GDSS embeds graphs into a continuous space and introduces Gaussian noise to the node features and the graph adjacency matrix. However, this approach removes the inherent sparsity of graphs, resulting in entirely noisy graphs where structural information, such as connectivity or cycle counts, remains undefined. Consequently, the continuous diffusion process can present challenges for the denoising network in capturing the structural properties of the data. In contrast, inspired by Discrete Denoising Diffusion Probabilistic Models (D3PMs)~\citep{austin2021structured}, a newly method DiGress~\citep{vignac2023digress} adopts a discrete diffusion process that systematically modifies graphs with noise. This process involves the addition or removal of edges and changes in categories. Additionally, DiGress employs a noise model that preserves the marginal distribution of node and edge types during the diffusion process. Furthermore, it introduces an innovative guidance procedure for conditioning graph generation on graph-level properties and enhances the input of the denoising network with auxiliary structural and spectral features. While DiGress has made notable advancements compared to GDSS, it still faces challenges in accurately estimating the joint distribution of measurements derived from node features and molecular graph structures. This challenge primarily arises from the approach of deriving separate embeddings for nodes and edges, treating them as distinct entities. Therefore, a more recent diffusion-based method called Wave-GD~\citep{cho2023multiresolution} has been introduced. Wave-GD harnesses the spectral dependencies between node and edge signals to more effectively characterize their joint distributions through a score-based diffusion model. By capturing their multi-resolution coherence, this model demonstrates the capability to generate high-fidelity molecules while preserving the frequency characteristics observed in the training samples.

\subsection{\textbf{Generation Strategy II: Fragment-based}}
Numerous methods have been proposed that utilize rational substructures, also known as fragments, as building blocks to generate high-quality molecules, which are categorized as ``\emph{fragment-based}" here. 

Among these methods, a considerable number are based on the variational autoencoder. An earlier typical work proposes a model named JT-VAE~\citep{jin2018junction} which first decomposes the molecular graph $G$ into its junction tree $\mathcal{T}$, where each node in the tree represents a substructure of the molecule. JT-VAE then encodes both the junction tree and molecular graph into their latent embeddings $\mathbf{z}_{\mathcal{T}}$ and $\mathbf{z}_{G}$, respectively. As for the decoding phase, JT-VAE first reconstructs junction tree from $\mathbf{z}_{\mathcal{T}}$ then generates molecule graph from the predicted junction tree by a graph decoder which learns how to assemble subgraphs. 

JT-VAE achieved a landmark 100\% validity, thanks to the implementation of a junction tree representation. This approach simplifies the process, as creating a tree structure is less complex compared to generating a general graph with degree constraints. Meanwhile, MHG-VAE~\citep{kajino2019molecular} refers to the scenario where the decoder produces invalid molecules as a \emph{decoding error issue}. While the tree representation effectively mitigates the \emph{decoding error issue}, it does come with certain limitations. This representation focuses only on connections at the fragment level and fails to define the specific atoms within the fragments that should be connected. Moreover, due to its lack of specificity at the atom level, information regarding stereochemistry is not retained. To compensate, all potential configurations are enumerated, and one is selected through the training of auxiliary neural networks. In contrast, MHG-VAE can address the \emph{decoding error issue} without the auxiliary neural networks. MHG-VAE's key innovation lies in its molecular hypergraph grammar (MHG), derived from the hyperedge replacement grammar (HRG)~\citep{drewes1997hyperedge}. In 
MHG, an atom is depicted as a hyperedge, while a bond is represented by a node. HRG generates a hypergraph by substituting a non-terminal hyperedge with another hypergraph. Integrating HRG with MHG allows for atom-level connections, and the stereochemistry can be seamlessly integrated into the grammar. Additionally, this approach respects the number of nodes linked to each hyperedge, which corresponds to the valency of atoms in molecular structures. Consequently, these principles facilitate the generation of valid molecules using a single VAE architecture.

HierVAE~\citep{jin2020hierarchical}, another VAE-based model designed by the same authors as JT-VAE, introduces larger and more flexible graph motifs as building blocks, exhibiting enhanced performance when dealing with larger molecules. The encoder of HierVAE generates a multi-resolution representation for each molecule, progressing in a fine-to-coarse fashion from atoms to connected motifs. Each hierarchical level in this model integrates the encoding of its lower-level constituents with the graph structure at that level. The autoregressive coarse-to-fine decoder of HierVAE adds motifs sequentially, one at a time. This process interweaves the selection of a new motif with the task of determining how it connects to the evolving molecular structure.

In the realm of drug discovery, there's often a necessity for a specific scaffold to be included in the synthesized molecule. MoLeR~\citep{maziarz2022learning}, also based on VAE, has been developed to cater to this need, facilitating the extension of partial molecules. MoLeR integrates motifs (molecule fragments) into its atom-by-atom generation process. When dealing with atoms that are part of a motif, MoLeR concatenates the initial chemically relevant features with the motif embedding. For atoms not associated with a motif, a special embedding vector is employed to indicate the absence of a motif. To deﬁne a concrete generation sequence, MoLeR initially opts for an initial atom selection. Subsequently, for each partial molecule, it proceeds to select the next atom from those that are adjacent to the atoms previously generated. Following each selection, in cases where the presently chosen atom is part of a motif, MoLeR incorporates the entire motif into the partial graph at once. Furthermore, MoLeR enhances its molecular generation process by integrating Molecular Swarm Optimization (MSO)~\citep{winter2019efficient}, which aids in the optimization of the molecular structures.

Previous methods like JT-VAE and HierVAE constructed the vocabulary of molecular fragments using simple hand-crafted rules, which may not effectively reveal frequent patterns in datasets. Besides, in previous \emph{fragment-based} methods, subgraph prediction and assembly are conducted either autoregressively, as in HierVAE, or according to a pre-defined tree structure, as in JT-VAE. However, both approaches have inherent limitations: each newly predicted subgraph can only attach to a local set of previously generated subgraphs, leading to inflexibility. To overcome these issues, PS-VAE~\citep{kong2022molecule} has been developed. PS-VAE begins by creating a vocabulary of molecular fragments from a given dataset, starting with distinct atoms and progressively merging neighboring fragments to update the vocabulary. This merge-and-update strategy leads to the formation of \emph{principal subgraphs}, a novel concept introduced in this paper, which represent frequent and significant repetitive patterns in molecules. PS-VAE also theoretically ensures that any \emph{principal subgraph} can be covered by the developed vocabulary. Additionally, PS-VAE introduces a two-step subgraph assembling strategy: it initially predicts a set of fragments sequentially and subsequently performs a global assembly of all generated subgraphs. This method reduces dependency on permutation and places more emphasis on global connectivity, offering a more robust approach compared to traditional fragment-based methods.

In the original paper of MiCaM~\citep{geng2023de}, another method based on VAE, the authors also highlight the crucial importance of developing an effective motif vocabulary. Accordingly, MiCaM devises an algorithm that identifies the most prevalent substructures based on their frequency within the molecule library. It scans the entire library to detect fragment pairs that frequently occur adjacent to each other within molecular graphs and subsequently merge these pairs into larger fragments. This merging process is repeated for a pre-determined number of steps, accumulating fragments to form a comprehensive motif vocabulary. The obtained motifs retain their structural connectivity information. Therefore, they are referred to as \emph{connection-aware motifs} in the paper. The generator of MiCaM operates by concurrently selecting motifs to be added and specifying their connection modes. During each step of the generation process, MiCaM concentrates on a nonterminal connection site within the current generated molecule. This site is then utilized to identify another connection, presenting two alternatives: (1) opting for a connection from the motif vocabulary, indicating the addition of a new motif, or (2) choosing a connection from the current molecular structure, a move that leads to the cyclization of the molecule.

Modof~\citep{chen2021deep} stands as another advanced deep generative method designed for molecule optimization. It operates by predicting a single disconnection site within a molecule and subsequently modifying the molecule by altering the fragments at that site, including elements such as ring systems, linkers, and side chains. What sets Modof apart from existing molecule optimization methods is its approach to learning from and encoding the disparities between molecules before and after optimization at a single disconnection site. When it comes to modifying a molecule, Modof generates only one fragment that represents the anticipated difference. This is achieved by decoding a sample obtained from the latent \emph{difference} space. Subsequently, Modof removes the original fragment at the disconnection site and replaces it with the newly generated fragment. Its model training also employs the VAE objective.

MoleculeChef~\citep{bradshaw2019model} has highlighted a significant issue in previous molecular generation methods. These earlier methods often failed to provide instructions on how to synthesize the molecules they generated. This lack of synthesis guidance implies that there are no assurances that the molecules produced by these methods can be practically synthesized in real-world laboratory scenarios. In contrast to prior research, MoleculeChef generates novel molecules by simulating virtual chemical reactions, closely mimicking the discovery process of molecules in the lab. n MoleculeChef, the encoder is responsible for mapping a multiset of reactants to a probability distribution in latent space, while the decoder generates this multiset of reactants sequentially using a RNN~\citep{medsker2001recurrent}. The resulting set of reactants is then processed through a reaction predictor to produce a final product. Notably, MoleculeChef employs the Molecular Transformer~\citep{schwaller2019molecular} as the chosen implementation for the reaction predictor. The paper acknowledges the possibility of utilizing a Variational Autoencoder (VAE) objective for training MoleculeChef. However, due to the inherent complexity of MoleculeChef's decoder, continuing to use the VAE objective would pose significant training challenges. As a solution, MoleculeChef ultimately opts to utilize the Wasserstein autoencoder (WAE)~\citep{tolstikhin2018wasserstein} objective. This decision is made to address the aforementioned training challenges associated with the VAE objective effectively.

There also exist several works using reinforcement learning (RL) to optimize the properties of generated molecules. An earlier work GCPN~\citep{you2018graph} stands out as a representative goal-directed 2D molecule generation approach. Compared with graph generative models struggling to incorporate desired molecular properties or constraints, RL excels at representing such hard constraints and desired properties by designing environment dynamics and reward functions. 
Besides, RL enables active exploration of the molecule space beyond the samples found in a dataset while the exploration capabilities of generative models are constrained by the limitations of the training dataset. GCPN formulates the graph generation task as a Markov Decision Process (MDP), where a molecule is constructed sequentially. This construction process involves either adding a bond to connect existing atoms within the graph or connecting a new fragment with the current molecular graph. The intermediate rewards include step-wise validity rewards and adversarial rewards, while the final rewards are computed as a sum over domain-specific rewards and adversarial rewards. The domain-specific rewards comprise the combination of final property scores, whereas the adversarial rewards are defined through a GAN. DeepGraphMolGen~\citep{khemchandani2020deepgraphmolgen} builds upon the foundation of GCPN and takes a further step in addressing the challenge of generating novel molecules with desired interaction properties. In DeepGraphMolGen, interaction binding models are learned from binding data using 
GCNs. Recognizing that experimentally obtained property scores may contain potentially significant errors, DeepGraphMolGen incorporates a robust loss for the model. Notably, in contrast to GCPN, the final reward in DeepGraphMolGen includes the pKi value~\citep{dubocovich1997melatonin} of the final molecule as predicted by the trained model.

While GCPN and DeepGraphMolGen are capable of generating molecules with desired properties, it remains challenging for them to generate molecules that simultaneously satisfy many property constraints. RationaleRL~\citep{jin2020multi}, an RL-based 2D molecule generation method, has been introduced to specifically tackle this limitation. Its core idea involves composing molecules from a vocabulary of substructures known as \emph{molecular rationales}.
The first step of RationaleRL is extracting rationales that are likely accountable for each property from molecules by Monte Carlo Tree Search (MCTS)~\citep{chaslot2010monte} and combining them for multiple properties. Specifically, during search process, each state in the search tree means a subgraph of the molecule and the property score of the subgraph indicates the reward. Then RationaleRL uses graph generative models to expand the rationales into full molecules. To generate realistic compounds, the graph generator is trained in two phases, namely pre-training phase and fine-tuning phase. After pre-training on a large set of real molecules, the graph generator is fine-tuned on property-specific rationales through multiple iterations using policy gradient.

Previous RL-based methods for goal-directed molecular design often emphasized relatively straightforward objectives, such as QED~\citep{ertl2009estimation}. However, achieving high scores in these simple molecular properties does not necessarily ensure drug-likeness or therapeutic potential, underscoring the importance of adopting more relevant design objectives in generative tasks. In contrast, another RL-based approach known as FREED~\citep{yang2021hit} places its focus on a more meaningful optimization target, namely, the docking score. This choice is made because docking simulations~\citep{schneidman2005patchdock,yan2017hdock} provide a more direct and practical proxy for assessing therapeutic potential, making it a valuable metric in the context of molecular design. FREED adopts a strategy for generating molecules that involves attaching a chemically realistic and pharmacochemically acceptable fragment unit to a given sub-graph at each step. Importantly, the model is enforced to create new bonds only at attachment sites that are considered suitable based on the fragment library preparation step. These strategies effectively leverage prior knowledge in medicinal chemistry, ensuring that molecule generation remains confined within the chemical space conducive to drug design. Additionally, FREED explores various explorative algorithms, incorporating curiosity-driven learning and prioritized experience replay (PER)~\citep{schaul2015prioritized}. In particular, FREED introduces an innovative PER method that defines priority based on the novelty of experiences, estimated by the predictive error or uncertainty of the auxiliary reward predictor's outcome. This approach is designed to mitigate the lack of robustness observed in previous methods and to encourage the exploration of diverse solutions during the molecular generation process.

Recently, a novel generative approach known as GFlowNet~\citep{bengio2021flow} has emerged, with close ties to RL. GFlowNet is designed to tackle the challenge of learning a stochastic policy for generating an object, such as a molecular graph, through a sequence of actions. The primary objective is to ensure that the probability of generating an object is directly proportional to a specified positive reward assigned to that object. GFlowNet views the generative process as a flow network, making it particularly adept at handling scenarios where various trajectories lead to the same final state. For instance, in molecular graph generation, there are multiple ways to sequentially add atoms to a molecule. In this context, GFlowNet conceptualizes the set of trajectories as a flow and transforms the flow consistency equations into a learning objective. This approach is analogous to how Bellman equations are utilized in Temporal Difference methods~\citep{sutton2018reinforcement}. Within the flow network framework, nodes represent states, and edges represent actions. GFlowNet finds practical application in molecule generation, where the `state' corresponds to the current molecule, and the `action' involves adding a fragment from a predefined vocabulary of fragments to the current molecule or terminating the generation process.

Another emerging line considers molecule generation as a sampling procedure. One noteworthy method in this category is MARS~\citep{xie2021mars}. The core concept of MARS involves initiating the process from a seed molecule and continually generating candidate molecules by making modifications to fragments of molecular graphs from previous iterations. In MARS, the task of molecular design is framed as an iterative editing process, with the overall objective being composed of multiple property scores. To search for optimal chemical compounds, MARS employs the annealed Markov Chain Monte Carlo (MCMC) sampling~\citep{geyer1992practical} technique. This approach allows for the exploration of chemicals with novel and diverse fragments. MARS utilizes graph neural networks (GNNs) to represent proposals for modifying molecular fragments. These GNNs adaptively learn their parameters to propose fragment modifications. While MPNNs are used in practice, other GNN architectures can also be integrated into the framework. Moreover, MARS leverages the sample paths generated on-the-fly to adaptively train the proposal network, eliminating the need for external annotated data. This adaptive learnable proposal mechanism enables MARS to continuously enhance the quality of molecule generation throughout the process.

MIMOSA is another molecule generation approach built on the MCMC sampling method. MIMOSA unfolds in three distinct stages. Initially, MIMOSA focuses on training a pair of property-neutral GNNs. These networks are tasked with the prediction of molecular topologies and substructure types. These substructures encompass atoms and rings. This step is crucial for improving the embeddings of molecules, aiding in the later stage of sampling. Subsequently, MIMOSA leverages these predictions to conduct three core substructure manipulations: \texttt{ADDITION}, \texttt{REPLACEMENT}, and \texttt{REMOVAL} to generate new molecule candidates. In its final stage, MIMOSA evaluates these newly generated molecules, assigning them weights based on criteria such as structural resemblance and drug property constraints. Molecules that fulfill these specified criteria are then chosen for additional processing rounds.

There's another fragment-based method called DEG~\citep{guo2022dataefficient}, which is also related to sampling. This method seeks to resolve two primary challenges. Existing methods predominantly rely on deep neural networks, necessitating extensive training on vast datasets, often comprising tens of thousands of examples. Contrarily, real-world scenarios usually present significantly smaller, class-specific chemical datasets, often just a few dozen samples, primarily due to the labor-intensive nature of experimentation and data acquisition. This limited dataset size presents a substantial challenge for deep learning-based generative models in effectively capturing the full scope of the molecular design landscape. In response, DEG emerges as a generative model optimized for data efficiency, capable of learning from much smaller datasets compared to standard benchmarks. Its core mechanism involves a learnable graph grammar, which generates molecules following a series of production rules. These rules are autonomously derived from the training data, requiring no manual intervention. Additionally, the model undergoes further refinement through grammar optimization, facilitating the integration of extra chemical insights.  Training of the DEG model is conducted through Monte Carlo (MC) sampling~\citep{metropolis1949monte} and the REINFORCE algorithm~\citep{williams1992simple}.

Additionally, it's noteworthy to mention a GAN-based method designed for molecular optimization, Mol-CycleGAN\citep{maziarka2020mol}. More specifically, it employs the CycleGAN~\citep{zhu2017unpaired} architecture. A key strength of Mol-CycleGAN lies in its capacity to discern and learn transformation rules from compound sets, based on their desired and undesired property values. It functions within a latent space, which is trained by another model. Specifically, in the case of Mol-CycleGAN, this latent space is derived from the JT-VAE model we discussed earlier. The capability of Mol-CycleGAN to generate molecules with particular desired properties is well-demonstrated, particularly in terms of structural and physicochemical attributes. Notably, the molecules produced by this model closely resemble their initial forms, with a tunable degree of similarity, adjustable through a designated hyperparameter.

\begin{table*}[tb!]
\centering
    \resizebox{0.96\linewidth}{!}{
\begin{tabular}{l|c|c|l}
\toprule
\textbf{Model}  & 
\textbf{Generation Strategy}

& \textbf{Methodology}& \textbf{Venue} \\
\midrule
\textbf{VGAE} & \emph{all-at-once} & VAE-based & NeurIPS workshop~\citeyear{kipf2016variational}\\
\textbf{GraphVAE} & \emph{all-at-once} & VAE-based  &ICANN~\citeyear{simonovsky2018graphvae} \\
\textbf{MPGVAE} & \emph{all-at-once} & VAE-based  &arXiv~\citeyear{flam2020graph} \\
\textbf{Regularized VAE} & \emph{all-at-once} & VAE-based & NeurIPS~\citeyear{ma2018constrained} \\
\textbf{MolGAN} & \emph{all-at-once} & GAN-based &ICML workshop~\citeyear{de2018molgan}\\
\textbf{GraphNVP} & \emph{all-at-once} & Flow-based  & arXiv~\citeyear{madhawa2019graphnvp} \\
{\textbf{GDSS}} & \emph{all-at-once} & Diffusion-based &  ICML~\citeyear{jo2022score}\\
{\textbf{DiGress}} & \emph{all-at-once} & Diffusion-based  & ICLR~\citeyear{vignac2023digress}\\
{\textbf{Wave-GD}}& \emph{all-at-once} & Diffusion-based & NeurIPS~\citeyear{cho2023multiresolution}\\
\midrule
\textbf{JT-VAE} & \emph{fragment-based} & VAE-based   & ICML~\citeyear{jin2018junction} \\
\textbf{MHGVAE} & \emph{fragment-based} & VAE-based  & ICML~\citeyear{kajino2019molecular} \\
\textbf{HierVAE} & \emph{fragment-based} & VAE-based  & ICML~\citeyear{jin2020hierarchical} \\
\textbf{MoleculeChef} & \emph{fragment-based} & VAE-based & NeurIPS~\citeyear{bradshaw2019model} \\
{\textbf{MoLeR}} & \emph{fragment-based} & VAE-based  & ICLR~\citeyear{maziarz2022learning} \\
{\textbf{PS-VAE}} & \emph{fragment-based} & VAE-based  & NeurIPS~\citeyear{kong2022molecule}\\
{\textbf{MiCaM}} & \emph{fragment-based} & VAE-based  & ICLR~\citeyear{geng2023de}\\
\textbf{Modof} & \emph{fragment-based} & VAE-based & Nature Machine Intelligence~\citeyear{chen2021deep} \\
\textbf{GCPN} & \emph{fragment-based} & RL-based & NeurIPS~\citeyear{you2018graph} \\
\textbf{DeepGraphMolGen} & \emph{fragment-based} & RL-based & Journal of Cheminformatics~\citeyear{khemchandani2020deepgraphmolgen} \\
\textbf{RationaleRL} & \emph{fragment-based} & RL-based & ICML~\citeyear{jin2020multi} \\
\textbf{FREED} & \emph{fragment-based} & RL-based & NeurIPS~\citeyear{yang2021hit} \\
\textbf{GFlowNet} & \emph{fragment-based} & GFlowNet-based  & NeurIPS~\citeyear{bengio2021flow} \\
\textbf{MARS} & \emph{fragment-based} & Sampling-based  & ICLR~\citeyear{xie2021mars} \\
\textbf{MIMOSA} & \emph{fragment-based} & Sampling-based  & AAAI~\citeyear{fu2021mimosa} \\
\textbf{DEG} & \emph{fragment-based} &  sampling-based & ICLR~\citeyear{guo2022dataefficient} \\
\textbf{Mol-CycleGAN} & \emph{fragment-based} & GAN-based  & Journal of Cheminformatics~\citeyear{maziarka2020mol} \\

\midrule
\textbf{CGVAE} & \emph{node-by-node} & VAE-based & NeurIPS~\citeyear{liu2018constrained} \\
\textbf{SbMolGen} & \emph{node-by-node} & VAE-based & Chemical Science~\citeyear{lim2020scaffold} \\
\textbf{GraphAF} & \emph{node-by-node} & Flow-based & ICLR~\citeyear{shi2020graphaf} \\
\textbf{GraphDF} & \emph{node-by-node} & Flow-based & ICML~\citeyear{luo2021graphdf} \\
\textbf{STGG} & \emph{node-by-node} & Spanning-tree-based & ICLR~\citeyear{ahn2022spanning} \\

\bottomrule
\end{tabular}}
\vspace{-6pt}
\caption{Recent representative works of 2D molecule generation.}
\label{tab:models}
\end{table*}

\subsection{\textbf{Generation Strategy III: Node-by-node}}

In addition to synthesizing molecules directly or employing substructures as foundational building blocks, recent advancements have introduced alternative methodologies for molecular generation. These novel approaches, often referred to as ``\emph{node-by-node}", involve constructing molecules at the most fundamental level, atom by atom.

CGVAE~\citep{liu2018constrained} is an VAE-based generative model which builds gated graph neural networks (GGNN)~\citep{li2015gated} into the encoder and decoder. CGVAE uses GGNN to embed each node in the input graph $G$ to a latent vector sampled from a diagonal normal distribution. The decoder of CGVAE initializes nodes with latent variables and generates edges between these nodes sequentially based on two decision functions: \texttt{FOCUS} and \texttt{EXPAND}. Specifically, the \texttt{FOCUS} function determines which node to visit and the \texttt{EXPAND} function decides which edges to add from the focus node in each step. The procedure will terminate when meeting the stop criteria. Notably, during the generation, all node representations should be updated once the generated subgraph changes. Furthermore, \texttt{EXPAND} function applies valency masking to guarantee chemical validity. CGVAE utilizes gradient ascent in the continuous latent space to optimize these molecules based on specific numerical properties.

\citeauthor{lim2020scaffold} propose another VAE-based method which is able to generate molecules with target properties while maintaining an arbitrary input scaffold as a substructure. In our survey, we refer to their method as SbMolGen. Its encoder adopts a variant of interaction network~\citep{battaglia2016interaction,gilmer2017neural} to encode one complete molecule graph $G$ into a latent vector $\mathbf{z}$, from which the decoder is trained to recover molecules. Specifically, the decoder takes a scaffold $S$ as input and sequentially adds nodes and edges to $S$ based on three loop stages namely \texttt{NODE\_ADDITION}, \texttt{EDGE\_ADDITION}, \texttt{NODE\_SELECTION} and a extra final stage named \texttt{ISOMER\_SELECTION}. Furthermore, we can concatenate the whole-molecule properties vector and scaffold properties vector with $\mathbf{z}$ sampled from latent space to condition the decoding process.

We have previously introduced two VAE-based methods for molecule generation in a node-by-node fashion. Additionally, there are flow-based methods available for generating molecules.
GraphAF~\citep{shi2020graphaf} is a representative flow-based model, whose concept is actually similar to the previously introduced GCPN. Both formulate the problem of molecular graph generation as a sequential decision process. Specifically, beginning with an empty graph, GraphAF sequentially generates a new node at each step, which is based on the structure of the current sub-graph. Subsequently, the edges connecting this newly added node with the existing ones are systematically formed, taking into account the existing graph structure. This iterative process continues until the generation of all nodes and edges is complete. The core concept of GraphAF involves defining an invertible transformation from a base distribution (like a multivariate Gaussian) to a molecular graph structure $G = (A, X)$. In every round of generation, GraphAF, given the existing sub-graph structure, employs a stack of multiple layers of a modified version of Relational GCN~\citep{schlichtkrull2018modeling} to derive the embeddings for each node. Following this, a sum-pooling operation is applied to these node embeddings to obtain the embedding of the entire sub-graph. This sub-graph embedding is then set as the mean and standard deviations of Gaussian distributions, which are subsequently used to generate the nodes and edges. Additionally, GraphAF suggests that the molecule generation process can be refined through reinforcement learning, aiming to optimize the properties of the molecules generated.

It's important to recognize that, akin to GraphNVP~\citep{madhawa2019graphnvp}, GraphAF also converts discrete graph data into continuous data through the addition of real-valued noise, a technique known as \emph{dequantization}. However, this \emph{dequantization} process hinders the models' ability to accurately represent the original discrete distribution of graph structures. This presents a significant challenge in model training, as it impedes the capability to accurately capture the true distribution of graph structures, resulting in a diverse range of molecules generated. To address the challenges posed by \emph{dequantization}, GraphDF~\citep{luo2021graphdf}, a novel approach building upon GraphAF, introduces the generation of molecular graphs using discrete latent variables. In GraphDF, all latent variables are discrete and sampled from multinomial distributions. We use a discrete ﬂow model to reversibly map discrete latent variables to new nodes and edges. The discrete transform used in the discrete ﬂow is a modulo shift transform. Apart from employing discrete latent variables, the process of molecule generation of GraphDF is similar to GraphAF.

\begin{table*}[tb!]
\centering
    \resizebox{0.96\linewidth}{!}{
\begin{tabular}{l|l|l|l}
\toprule
\textbf{Dataset}  & \textbf{Description} & \textbf{Number of molecules} & \textbf{Link}\\
\midrule
QM9& \makecell[l]{Stable small organic molecules made up of CHONF atoms} & $133,885$ & \href{http://quantum-machine.org/datasets/}{http://quantum-machine.org/datasets/} \\
\midrule
GDB-17& \makecell[l]{Enumeration of small organic molecules up to 17 atoms} & $>166,000,000,000$ & \href{http://gdb.unibe.ch/downloads/}{http://gdb.unibe.ch/downloads/} \\
\midrule
ZINC15& \makecell[l]{Commercially available compounds} & $>750,000,000$ & \href{http://zinc15.docking.org/}{http://zinc15.docking.org/}\\
\midrule
ChEMBL& \makecell[l]{Bioactive molecules with drug-like properties} & $>2,000,000$ & \href{https://www.ebi.ac.uk/chembl/}{https://www.ebi.ac.uk/chembl/}\\
\midrule
PubChemQC& \makecell[l]{Compounds with quantum chemistry estimated \\property based on density functional theory} & $3,981,230$ & \href{http://pubchemqc.riken.jp/}{http://pubchemqc.riken.jp/}\\
\midrule
DrugBank& \makecell[l]{FDA-approved drugs and other drugs public available} & $>14,000$ & \href{https://www.drugbank.ca/}{https://www.drugbank.ca/}\\
\bottomrule
\end{tabular}}
\vspace{-6pt}
\caption{Representative datasets for 2D molecule generation.}
\label{tab:datasets}
\end{table*}

STGG~\citep{ahn2022spanning} stands out from prior VAE-based and flow-based methods as the first framework to utilize a spanning tree-based approach for molecular graph generation. This unique methodology conceptualizes the generation of molecular graphs through the construction of a spanning tree along with residual edges. This approach leverages the inherent sparsity found in molecular graphs, enabling the use of efficient tree-constructive operations to establish molecular graph connectivity. STGG ensures that the generated molecular graphs adhere to chemical valence rules by applying constraints based on the intermediate graph structure formed during the construction process. Additionally, STGG introduces an innovative Transformer~\citep{vaswani2017attention} architecture, incorporating tree-based relative positional encodings, to effectively facilitate the tree construction procedure.

\subsection{Discussion}
The generation at different levels can have different advantages in specific applications. While in general, the fine-grained level manipulation at node level is flexible, while it may be less efficient for generation and has difficulty in modeling higher-level (sub)structure information. While the fragment-based pipeline allows edition of sub-structures of a molecule, which can be often meaningful to some specific functionality and reaction. Finally, the once-for-all scheme can be efficient while it may sometimes lack enough flexibility for incremental generation and optimization.

\section{Datasets and Evaluation Metrics}
\label{sec:data}

\paragraph{\textbf{Datasets}} We present a compilation of prominent publicly accessible datasets, commonly employed in molecule generation and optimization endeavors, as depicted in Table~\ref{tab:datasets}. Among them, ChEMBL and DrugBank are continually evolving, with periodic updates to their contents.

\paragraph{\textbf{Evaluation Metrics}} Generation and optimization for molecules adopt two different sets of evaluation metrics. Molecule generation evaluates the overall quality of generated molecules from a statistical perspective in terms of these metrics, including \emph{validity} (the percentage of generated molecules that are chemically valid), \emph{novelty} (the fraction of generated molecules not appearing in the training data), \emph{diversity} (the pairwise molecular distance among generated molecules), \emph{uniqueness} (ratio of unique molecules) and \emph{reconstruction} (the percentage of molecules which can be reconstructed from their latent variables). For molecule optimization, it adopts another set of metrics for evaluation on the basis of multi-property of generated molecules , such as quantitative estimate of drug-likeness (QED)~\citep{ertl2009estimation}, synthetic accessibility (SA)~\citep{bickerton2012quantifying}, {octanol-water partition coefficients} (logP)~\citep{sangster1997octanol} and so on.

\section{Challenges and Future Directions}
\label{sec:challenge}
Despite the significant achievements of graph-based deep learning in automating molecule design, the complexity of molecular structures presents ongoing challenges. In this section, we suggest three future directions for further research.
\paragraph{\textbf{Macro-molecules Design}}
The current body of research primarily concentrates on the design of small molecules, and its effectiveness diminishes considerably when applied to the design of larger molecules such as polymers. The complexity of large molecular systems is inherently much greater than that of small molecular systems, naturally leading to increased modeling difficulties. The failure also stems from the many generation steps required to realize larger molecules and the associated challenges with gradients across the iterative steps~\citep{jin2020hierarchical}. Therefore, it is imperative to devise novel approaches specifically tailored to handle larger molecules.

\paragraph{\textbf{3D Drug Discovery}} 
The generation of 3D molecular geometries is an area that has not been extensively explored. Compared to 1D SMILES-based and 2D graph-based representations, the addition of a third dimension considerably broadens the molecular space to be examined, thereby raising the level of complexity~\citep{yang2021knowledge}. Nonetheless, the generation of 3D molecules is both meaningful and necessary, given that accurate 3D coordinates are crucial for precise prediction of quantum properties~\citep{schutt2017schnet}. Despite its importance, there are relatively few studies focused on this aspect, highlighting a need for further research and development in this field.

\paragraph{\textbf{Structure-based Drug Design}} 
Chemical space is vast, yet the subset of molecules with certain desirable properties is much smaller by contrast, e.g. activity against a given target, that makes them well suited for the discovery of drug candidates~\citep{zhavoronkov2019deep}. Structure-based drug discovery aims to design small-molecule ligands that bind with high affinity and specificity to pre-determined protein targets~\citep{anderson2003process}, which is a fundamental and challenging task in drug discovery, essentially sampling compounds from promising sub-regions of chemical space. Nevertheless, the successful application of machine learning to the problem of structure-based drug design remains relatively limited in the current literature. This highlights a substantial challenge and, simultaneously, presents a significant opportunity for breakthroughs and progress in future research within this domain.

\section{Conclusion}
\label{sec:conclusion}

The generation of molecules with desired properties holds paramount importance, particularly in the pharmaceutical industry. In this context, we have introduced a broad array of graph-based deep learning models for molecular deisgn, categorizing them into three distinct groups based on their generative strategies. Additionally, we have compiled a comprehensive overview of public datasets and the evaluation metrics widely used in this field. In the end, we delve into the challenges and prospective future developments in this dynamic and evolving area.

\section*{Acknowledgments}
This work was in part supported by National Key Research and Development
Program of China (2020AAA0107600), and NSFC (62222607).

\bibliographystyle{plainnat}
\bibliography{ref.bib}

\begin{thebibliography}{88}
\providecommand{\natexlab}[1]{#1}
\providecommand{\url}[1]{\texttt{#1}}
\expandafter\ifx\csname urlstyle\endcsname\relax
  \providecommand{\doi}[1]{doi: #1}\else
  \providecommand{\doi}{doi: \begingroup \urlstyle{rm}\Url}\fi

\bibitem[Ahn et~al.(2022)Ahn, Chen, Wang, and Song]{ahn2022spanning}
Sungsoo Ahn, Binghong Chen, Tianzhe Wang, and Le~Song.
\newblock Spanning tree-based graph generation for molecules.
\newblock In \emph{International Conference on Learning Representations}, 2022.
\newblock URL \url{https://openreview.net/forum?id=w60btE_8T2m}.

\bibitem[Alshehri et~al.(2020)Alshehri, Gani, and You]{alshehri2020deep}
Abdulelah~S Alshehri, Rafiqul Gani, and Fengqi You.
\newblock Deep learning and knowledge-based methods for computer-aided molecular design—toward a unified approach: State-of-the-art and future directions.
\newblock \emph{Computers \& Chemical Engineering}, 2020.

\bibitem[Anderson(2003)]{anderson2003process}
Amy~C Anderson.
\newblock The process of structure-based drug design.
\newblock \emph{Chemistry \& biology}, 10\penalty0 (9):\penalty0 787--797, 2003.

\bibitem[Austin et~al.(2021)Austin, Johnson, Ho, Tarlow, and Van Den~Berg]{austin2021structured}
Jacob Austin, Daniel~D Johnson, Jonathan Ho, Daniel Tarlow, and Rianne Van Den~Berg.
\newblock Structured denoising diffusion models in discrete state-spaces.
\newblock \emph{Advances in Neural Information Processing Systems}, 34:\penalty0 17981--17993, 2021.

\bibitem[Battaglia et~al.(2016)Battaglia, Pascanu, Lai, Jimenez~Rezende, et~al.]{battaglia2016interaction}
Peter Battaglia, Razvan Pascanu, Matthew Lai, Danilo Jimenez~Rezende, et~al.
\newblock Interaction networks for learning about objects, relations and physics.
\newblock \emph{Advances in neural information processing systems}, 29, 2016.

\bibitem[Bengio et~al.(2021)Bengio, Jain, Korablyov, Precup, and Bengio]{bengio2021flow}
Emmanuel Bengio, Moksh Jain, Maksym Korablyov, Doina Precup, and Yoshua Bengio.
\newblock Flow network based generative models for non-iterative diverse candidate generation.
\newblock \emph{Advances in Neural Information Processing Systems}, 34:\penalty0 27381--27394, 2021.

\bibitem[Bickerton et~al.(2012)Bickerton, Paolini, Besnard, Muresan, and Hopkins]{bickerton2012quantifying}
G~Richard Bickerton, Gaia~V Paolini, J{\'e}r{\'e}my Besnard, Sorel Muresan, and Andrew~L Hopkins.
\newblock Quantifying the chemical beauty of drugs.
\newblock \emph{Nature chemistry}, 2012.

\bibitem[Bradshaw et~al.(2019)Bradshaw, Paige, Kusner, Segler, and Hern{\'a}ndez-Lobato]{bradshaw2019model}
John Bradshaw, Brooks Paige, Matt~J Kusner, Marwin Segler, and Jos{\'e}~Miguel Hern{\'a}ndez-Lobato.
\newblock A model to search for synthesizable molecules.
\newblock \emph{Advances in Neural Information Processing Systems}, 32, 2019.

\bibitem[Chaslot(2010)]{chaslot2010monte}
Guillaume Maurice Jean-Bernard~Chaslot Chaslot.
\newblock \emph{Monte-carlo tree search}.
\newblock Maastricht University, 2010.

\bibitem[Chen et~al.(2021)Chen, Min, Parthasarathy, and Ning]{chen2021deep}
Ziqi Chen, Martin~Renqiang Min, Srinivasan Parthasarathy, and Xia Ning.
\newblock A deep generative model for molecule optimization via one fragment modification.
\newblock \emph{Nature machine intelligence}, 3\penalty0 (12):\penalty0 1040--1049, 2021.

\bibitem[Cho et~al.(2023)Cho, Jeong, Jeon, Ahn, and Kim]{cho2023multiresolution}
Hyuna Cho, Minjae Jeong, Sooyeon Jeon, Sungsoo Ahn, and Won~Hwa Kim.
\newblock Multi-resolution spectral coherence for graph generation with score-based diffusion.
\newblock In \emph{Thirty-seventh Conference on Neural Information Processing Systems}, 2023.
\newblock URL \url{https://openreview.net/forum?id=qUlpDjYnsp}.

\bibitem[De~Cao and Kipf(2018)]{de2018molgan}
Nicola De~Cao and Thomas Kipf.
\newblock {MolGAN: An implicit generative model for small molecular graphs}.
\newblock \emph{ICML 2018 workshop on Theoretical Foundations and Applications of Deep Generative Models}, 2018.

\bibitem[Dinh et~al.(2017)Dinh, Sohl-Dickstein, and Bengio]{dinh2017density}
Laurent Dinh, Jascha Sohl-Dickstein, and Samy Bengio.
\newblock Density estimation using real {NVP}.
\newblock In \emph{International Conference on Learning Representations}, 2017.
\newblock URL \url{https://openreview.net/forum?id=HkpbnH9lx}.

\bibitem[Drewes et~al.(1997)Drewes, Kreowski, and Habel]{drewes1997hyperedge}
Frank Drewes, H-J Kreowski, and Annegret Habel.
\newblock Hyperedge replacement graph grammars.
\newblock In \emph{Handbook Of Graph Grammars And Computing By Graph Transformation: Volume 1: Foundations}, pages 95--162. World Scientific, 1997.

\bibitem[Dubocovich et~al.(1997)Dubocovich, Masana, Iacob, and Sauri]{dubocovich1997melatonin}
Margarita~L Dubocovich, Monica~I Masana, Stanca Iacob, and Daniel~M Sauri.
\newblock Melatonin receptor antagonists that differentiate between the human mel1a and mel1b recombinant subtypes are used to assess the pharmacological profile of the rabbit retina ml1 presynaptic heteroreceptor.
\newblock \emph{Naunyn-Schmiedeberg's archives of pharmacology}, 355:\penalty0 365--375, 1997.

\bibitem[Elton et~al.(2019)Elton, Boukouvalas, Fuge, and Chung]{elton2019deep}
Daniel~C Elton, Zois Boukouvalas, Mark~D Fuge, and Peter~W Chung.
\newblock Deep learning for molecular design—a review of the state of the art.
\newblock \emph{Molecular Systems Design \& Engineering}, 2019.

\bibitem[Ertl and Schuffenhauer(2009)]{ertl2009estimation}
Peter Ertl and Ansgar Schuffenhauer.
\newblock Estimation of synthetic accessibility score of drug-like molecules based on molecular complexity and fragment contributions.
\newblock \emph{Journal of cheminformatics}, 2009.

\bibitem[Faez et~al.(2021)Faez, Ommi, Baghshah, and Rabiee]{faez2021deep}
Faezeh Faez, Yassaman Ommi, Mahdieh~Soleymani Baghshah, and Hamid~R Rabiee.
\newblock Deep graph generators: A survey.
\newblock \emph{IEEE Access}, 2021.

\bibitem[Flam-Shepherd et~al.(2020)Flam-Shepherd, Wu, and Aspuru-Guzik]{flam2020graph}
Daniel Flam-Shepherd, Tony Wu, and Alan Aspuru-Guzik.
\newblock Graph deconvolutional generation.
\newblock \emph{arXiv preprint arXiv:2002.07087}, 2020.

\bibitem[Fu et~al.(2021)Fu, Xiao, Li, Glass, and Sun]{fu2021mimosa}
Tianfan Fu, Cao Xiao, Xinhao Li, Lucas~M Glass, and Jimeng Sun.
\newblock Mimosa: Multi-constraint molecule sampling for molecule optimization.
\newblock In \emph{Proceedings of the AAAI Conference on Artificial Intelligence}, volume~35, pages 125--133, 2021.

\bibitem[Garcia~Satorras et~al.(2021)Garcia~Satorras, Hoogeboom, Fuchs, Posner, and Welling]{garcia2021n}
Victor Garcia~Satorras, Emiel Hoogeboom, Fabian Fuchs, Ingmar Posner, and Max Welling.
\newblock E(n) equivariant normalizing flows.
\newblock \emph{Advances in Neural Information Processing Systems}, 34:\penalty0 4181--4192, 2021.

\bibitem[Gebauer et~al.(2019)Gebauer, Gastegger, and Sch{\"u}tt]{gebauer2019symmetry}
Niklas Gebauer, Michael Gastegger, and Kristof Sch{\"u}tt.
\newblock Symmetry-adapted generation of 3d point sets for the targeted discovery of molecules.
\newblock \emph{Advances in neural information processing systems}, 32, 2019.

\bibitem[Geng et~al.(2023)Geng, Xie, Xia, Wu, Qin, Wang, Zhang, Wu, and Liu]{geng2023de}
Zijie Geng, Shufang Xie, Yingce Xia, Lijun Wu, Tao Qin, Jie Wang, Yongdong Zhang, Feng Wu, and Tie-Yan Liu.
\newblock De novo molecular generation via connection-aware motif mining.
\newblock In \emph{The Eleventh International Conference on Learning Representations}, 2023.
\newblock URL \url{https://openreview.net/forum?id=Q_Jexl8-qDi}.

\bibitem[Geyer(1992)]{geyer1992practical}
Charles~J Geyer.
\newblock Practical markov chain monte carlo.
\newblock \emph{Statistical science}, 1992.

\bibitem[Gilmer et~al.(2017)Gilmer, Schoenholz, Riley, Vinyals, and Dahl]{gilmer2017neural}
Justin Gilmer, Samuel~S Schoenholz, Patrick~F Riley, Oriol Vinyals, and George~E Dahl.
\newblock Neural message passing for quantum chemistry.
\newblock In \emph{International Conference on Machine Learning}, pages 1263--1272. PMLR, 2017.

\bibitem[G{\'o}mez-Bombarelli et~al.(2018)G{\'o}mez-Bombarelli, Wei, Duvenaud, Hern{\'a}ndez-Lobato, S{\'a}nchez-Lengeling, Sheberla, Aguilera-Iparraguirre, Hirzel, Adams, and Aspuru-Guzik]{gomez2018automatic}
Rafael G{\'o}mez-Bombarelli, Jennifer~N Wei, David Duvenaud, Jos{\'e}~Miguel Hern{\'a}ndez-Lobato, Benjam{\'\i}n S{\'a}nchez-Lengeling, Dennis Sheberla, Jorge Aguilera-Iparraguirre, Timothy~D Hirzel, Ryan~P Adams, and Al{\'a}n Aspuru-Guzik.
\newblock Automatic chemical design using a data-driven continuous representation of molecules.
\newblock \emph{ACS central science}, 2018.

\bibitem[Goodfellow et~al.(2014)Goodfellow, Pouget-Abadie, Mirza, Xu, Warde-Farley, Ozair, Courville, and Bengio]{goodfellow2014generative}
Ian Goodfellow, Jean Pouget-Abadie, Mehdi Mirza, Bing Xu, David Warde-Farley, Sherjil Ozair, Aaron Courville, and Yoshua Bengio.
\newblock Generative adversarial nets.
\newblock \emph{Advances in neural information processing systems}, 27, 2014.

\bibitem[Griffiths and Hern{\'a}ndez-Lobato(2020)]{griffiths2020constrained}
Ryan-Rhys Griffiths and Jos{\'e}~Miguel Hern{\'a}ndez-Lobato.
\newblock Constrained bayesian optimization for automatic chemical design using variational autoencoders.
\newblock \emph{Chemical science}, 2020.

\bibitem[Grisoni et~al.(2020)Grisoni, Moret, Lingwood, and Schneider]{grisoni2020bidirectional}
Francesca Grisoni, Michael Moret, Robin Lingwood, and Gisbert Schneider.
\newblock Bidirectional molecule generation with recurrent neural networks.
\newblock \emph{Journal of chemical information and modeling}, 2020.

\bibitem[Guo et~al.(2022)Guo, Thost, Li, Das, Chen, and Matusik]{guo2022dataefficient}
Minghao Guo, Veronika Thost, Beichen Li, Payel Das, Jie Chen, and Wojciech Matusik.
\newblock Data-efficient graph grammar learning for molecular generation.
\newblock In \emph{International Conference on Learning Representations}, 2022.
\newblock URL \url{https://openreview.net/forum?id=l4IHywGq6a}.

\bibitem[Guo and Zhao(2022)]{guo2022systematic}
Xiaojie Guo and Liang Zhao.
\newblock A systematic survey on deep generative models for graph generation.
\newblock \emph{IEEE Transactions on Pattern Analysis and Machine Intelligence}, 45\penalty0 (5):\penalty0 5370--5390, 2022.

\bibitem[Guo et~al.(2023)Guo, Guo, Nan, Tian, Iyer, Ma, Wiest, Zhang, Wang, Zhang, and Chawla]{guo2023survey}
Zhichun Guo, Kehan Guo, Bozhao Nan, Yijun Tian, Roshni~G. Iyer, Yihong Ma, Olaf Wiest, Xiangliang Zhang, Wei Wang, Chuxu Zhang, and Nitesh~V. Chawla.
\newblock Graph-based molecular representation learning.
\newblock In \emph{Proceedings of the Thirty-Second International Joint Conference on Artificial Intelligence, {IJCAI-23}}, pages 6638--6646, 2023.

\bibitem[Ho et~al.(2020)Ho, Jain, and Abbeel]{ho2020denoising}
Jonathan Ho, Ajay Jain, and Pieter Abbeel.
\newblock Denoising diffusion probabilistic models.
\newblock \emph{Advances in neural information processing systems}, 33:\penalty0 6840--6851, 2020.

\bibitem[Jim{\'e}nez-Luna et~al.(2020)Jim{\'e}nez-Luna, Grisoni, and Schneider]{jimenez2020drug}
Jos{\'e} Jim{\'e}nez-Luna, Francesca Grisoni, and Gisbert Schneider.
\newblock Drug discovery with explainable artificial intelligence.
\newblock \emph{Nature Machine Intelligence}, 2020.

\bibitem[Jin et~al.(2018)Jin, Barzilay, and Jaakkola]{jin2018junction}
Wengong Jin, Regina Barzilay, and Tommi Jaakkola.
\newblock Junction tree variational autoencoder for molecular graph generation.
\newblock In \emph{International Conference on Machine Learning}, pages 2323--2332. PMLR, 2018.

\bibitem[Jin et~al.(2020{\natexlab{a}})Jin, Barzilay, and Jaakkola]{jin2020hierarchical}
Wengong Jin, Regina Barzilay, and Tommi Jaakkola.
\newblock Hierarchical generation of molecular graphs using structural motifs.
\newblock In \emph{International Conference on Machine Learning}, pages 4839--4848. PMLR, 2020{\natexlab{a}}.

\bibitem[Jin et~al.(2020{\natexlab{b}})Jin, Barzilay, and Jaakkola]{jin2020multi}
Wengong Jin, Regina Barzilay, and Tommi Jaakkola.
\newblock Multi-objective molecule generation using interpretable substructures.
\newblock In \emph{International Conference on Machine Learning}, pages 4849--4859. PMLR, 2020{\natexlab{b}}.

\bibitem[Jo et~al.(2022)Jo, Lee, and Hwang]{jo2022score}
Jaehyeong Jo, Seul Lee, and Sung~Ju Hwang.
\newblock Score-based generative modeling of graphs via the system of stochastic differential equations.
\newblock In \emph{International Conference on Machine Learning}, pages 10362--10383. PMLR, 2022.

\bibitem[Kaelbling et~al.(1996)Kaelbling, Littman, and Moore]{kaelbling1996reinforcement}
Leslie~Pack Kaelbling, Michael~L Littman, and Andrew~W Moore.
\newblock Reinforcement learning: A survey.
\newblock \emph{Journal of artificial intelligence research}, 4:\penalty0 237--285, 1996.

\bibitem[Kajino(2019)]{kajino2019molecular}
Hiroshi Kajino.
\newblock Molecular hypergraph grammar with its application to molecular optimization.
\newblock In \emph{International Conference on Machine Learning}, pages 3183--3191. PMLR, 2019.

\bibitem[Kang and Cho(2018)]{kang2018conditional}
Seokho Kang and Kyunghyun Cho.
\newblock Conditional molecular design with deep generative models.
\newblock \emph{Journal of chemical information and modeling}, 2018.

\bibitem[Khemchandani et~al.(2020)Khemchandani, O’Hagan, Samanta, Swainston, Roberts, Bollegala, and Kell]{khemchandani2020deepgraphmolgen}
Yash Khemchandani, Stephen O’Hagan, Soumitra Samanta, Neil Swainston, Timothy~J Roberts, Danushka Bollegala, and Douglas~B Kell.
\newblock Deepgraphmolgen, a multi-objective, computational strategy for generating molecules with desirable properties: a graph convolution and reinforcement learning approach.
\newblock \emph{Journal of Cheminformatics}, 2020.

\bibitem[Kim et~al.(2020)Kim, Kim, Lee, Bae, Park, and Nam]{kim2020artificial}
Hyunho Kim, Eunyoung Kim, Ingoo Lee, Bongsung Bae, Minsu Park, and Hojung Nam.
\newblock Artificial intelligence in drug discovery: a comprehensive review of data-driven and machine learning approaches.
\newblock \emph{Biotechnology and Bioprocess Engineering}, 2020.

\bibitem[Kingma and Welling(2014)]{kingma2014auto}
Diederik~P Kingma and Max Welling.
\newblock Auto-encoding variational bayes.
\newblock \emph{stat}, 2014.

\bibitem[Kingma and Dhariwal(2018)]{kingma2018glow}
Durk~P Kingma and Prafulla Dhariwal.
\newblock Glow: Generative flow with invertible 1x1 convolutions.
\newblock \emph{Advances in neural information processing systems}, 31, 2018.

\bibitem[Kipf and Welling(2016)]{kipf2016variational}
Thomas~N Kipf and Max Welling.
\newblock Variational graph auto-encoders.
\newblock \emph{NIPS Workshop on Bayesian Deep Learning}, 2016.

\bibitem[Kong et~al.(2022)Kong, Huang, Tan, and Liu]{kong2022molecule}
Xiangzhe Kong, Wenbing Huang, Zhixing Tan, and Yang Liu.
\newblock Molecule generation by principal subgraph mining and assembling.
\newblock \emph{Advances in Neural Information Processing Systems}, 35:\penalty0 2550--2563, 2022.

\bibitem[Li et~al.(2016)Li, Tarlow, Brockschmidt, and Zemel]{li2015gated}
Yujia Li, Daniel Tarlow, Marc Brockschmidt, and Richard~S. Zemel.
\newblock Gated graph sequence neural networks.
\newblock In Yoshua Bengio and Yann LeCun, editors, \emph{4th International Conference on Learning Representations, {ICLR} 2016, San Juan, Puerto Rico, May 2-4, 2016, Conference Track Proceedings}, 2016.
\newblock URL \url{http://arxiv.org/abs/1511.05493}.

\bibitem[Lim et~al.(2020)Lim, Hwang, Moon, Kim, and Kim]{lim2020scaffold}
Jaechang Lim, Sang-Yeon Hwang, Seokhyun Moon, Seungsu Kim, and Woo~Youn Kim.
\newblock Scaffold-based molecular design with a graph generative model.
\newblock \emph{Chemical science}, 2020.

\bibitem[Liu et~al.(2018)Liu, Allamanis, Brockschmidt, and Gaunt]{liu2018constrained}
Qi~Liu, Miltiadis Allamanis, Marc Brockschmidt, and Alexander Gaunt.
\newblock Constrained graph variational autoencoders for molecule design.
\newblock \emph{Advances in neural information processing systems}, 31, 2018.

\bibitem[Luo and Ji(2022)]{luo2022an}
Youzhi Luo and Shuiwang Ji.
\newblock An autoregressive flow model for 3d molecular geometry generation from scratch.
\newblock In \emph{International Conference on Learning Representations}, 2022.
\newblock URL \url{https://openreview.net/forum?id=C03Ajc-NS5W}.

\bibitem[Luo et~al.(2021)Luo, Yan, and Ji]{luo2021graphdf}
Youzhi Luo, Keqiang Yan, and Shuiwang Ji.
\newblock Graphdf: A discrete flow model for molecular graph generation.
\newblock In \emph{International Conference on Machine Learning}, pages 7192--7203. PMLR, 2021.

\bibitem[Ma et~al.(2018)Ma, Chen, and Xiao]{ma2018constrained}
Tengfei Ma, Jie Chen, and Cao Xiao.
\newblock Constrained generation of semantically valid graphs via regularizing variational autoencoders.
\newblock \emph{Advances in Neural Information Processing Systems}, 31, 2018.

\bibitem[Madhawa et~al.(2019)Madhawa, Ishiguro, Nakago, and Abe]{madhawa2019graphnvp}
Kaushalya Madhawa, Katushiko Ishiguro, Kosuke Nakago, and Motoki Abe.
\newblock Graphnvp: An invertible flow model for generating molecular graphs.
\newblock \emph{arXiv:1905.11600}, 2019.

\bibitem[Maziarka et~al.(2020)Maziarka, Pocha, Kaczmarczyk, Rataj, Danel, and Warcho{\l}]{maziarka2020mol}
{\L}ukasz Maziarka, Agnieszka Pocha, Jan Kaczmarczyk, Krzysztof Rataj, Tomasz Danel, and Micha{\l} Warcho{\l}.
\newblock Mol-cyclegan: a generative model for molecular optimization.
\newblock \emph{Journal of Cheminformatics}, 2020.

\bibitem[Maziarz et~al.(2022)Maziarz, Jackson-Flux, Cameron, Sirockin, Schneider, Stiefl, Segler, and Brockschmidt]{maziarz2022learning}
Krzysztof Maziarz, Henry~Richard Jackson-Flux, Pashmina Cameron, Finton Sirockin, Nadine Schneider, Nikolaus Stiefl, Marwin Segler, and Marc Brockschmidt.
\newblock Learning to extend molecular scaffolds with structural motifs.
\newblock In \emph{International Conference on Learning Representations}, 2022.
\newblock URL \url{https://openreview.net/forum?id=ZTsoE8G3GG}.

\bibitem[Medsker and Jain(2001)]{medsker2001recurrent}
Larry~R Medsker and LC~Jain.
\newblock Recurrent neural networks.
\newblock \emph{Design and Applications}, 5\penalty0 (64-67):\penalty0 2, 2001.

\bibitem[Metropolis and Ulam(1949)]{metropolis1949monte}
Nicholas Metropolis and Stanislaw Ulam.
\newblock The monte carlo method.
\newblock \emph{Journal of the American statistical association}, 44\penalty0 (247):\penalty0 335--341, 1949.

\bibitem[Putin et~al.(2018)Putin, Asadulaev, Vanhaelen, Ivanenkov, Aladinskaya, Aliper, and Zhavoronkov]{putin2018adversarial}
Evgeny Putin, Arip Asadulaev, Quentin Vanhaelen, Yan Ivanenkov, Anastasia~V Aladinskaya, Alex Aliper, and Alex Zhavoronkov.
\newblock Adversarial threshold neural computer for molecular de novo design.
\newblock \emph{Molecular pharmaceutics}, 2018.

\bibitem[Redkar et~al.(2020)Redkar, Mondal, Joseph, and Hareesha]{redkar2020machine}
Shweta Redkar, Sukanta Mondal, Alex Joseph, and KS~Hareesha.
\newblock A machine learning approach for drug-target interaction prediction using wrapper feature selection and class balancing.
\newblock \emph{Molecular informatics}, 2020.

\bibitem[Roney et~al.(2022)Roney, Maragakis, Skopp, and Shaw]{roney2022generating}
James~P. Roney, Paul Maragakis, Peter Skopp, and David~E. Shaw.
\newblock Generating realistic 3d molecules with an equivariant conditional likelihood model, 2022.
\newblock URL \url{https://openreview.net/forum?id=Snqhqz4LdK}.

\bibitem[Sangster(1997)]{sangster1997octanol}
James~M Sangster.
\newblock \emph{Octanol-water partition coefficients: fundamentals and physical chemistry}, volume~1.
\newblock John Wiley \& Sons, 1997.

\bibitem[Schaul et~al.(2015)Schaul, Quan, Antonoglou, and Silver]{schaul2015prioritized}
Tom Schaul, John Quan, Ioannis Antonoglou, and David Silver.
\newblock Prioritized experience replay.
\newblock \emph{arXiv preprint arXiv:1511.05952}, 2015.

\bibitem[Schlichtkrull et~al.(2018)Schlichtkrull, Kipf, Bloem, Van Den~Berg, Titov, and Welling]{schlichtkrull2018modeling}
Michael Schlichtkrull, Thomas~N Kipf, Peter Bloem, Rianne Van Den~Berg, Ivan Titov, and Max Welling.
\newblock Modeling relational data with graph convolutional networks.
\newblock In \emph{The Semantic Web: 15th International Conference, ESWC 2018, Heraklion, Crete, Greece, June 3--7, 2018, Proceedings 15}, pages 593--607. Springer, 2018.

\bibitem[Schneidman-Duhovny et~al.(2005)Schneidman-Duhovny, Inbar, Nussinov, and Wolfson]{schneidman2005patchdock}
Dina Schneidman-Duhovny, Yuval Inbar, Ruth Nussinov, and Haim~J Wolfson.
\newblock Patchdock and symmdock: servers for rigid and symmetric docking.
\newblock \emph{Nucleic acids research}, 33\penalty0 (suppl\_2):\penalty0 W363--W367, 2005.

\bibitem[Sch{\"u}tt et~al.(2017)Sch{\"u}tt, Kindermans, Sauceda~Felix, Chmiela, Tkatchenko, and M{\"u}ller]{schutt2017schnet}
Kristof Sch{\"u}tt, Pieter-Jan Kindermans, Huziel~Enoc Sauceda~Felix, Stefan Chmiela, Alexandre Tkatchenko, and Klaus-Robert M{\"u}ller.
\newblock Schnet: A continuous-filter convolutional neural network for modeling quantum interactions.
\newblock \emph{Advances in neural information processing systems}, 30, 2017.

\bibitem[Schwaller et~al.(2019)Schwaller, Laino, Gaudin, Bolgar, Hunter, Bekas, and Lee]{schwaller2019molecular}
Philippe Schwaller, Teodoro Laino, Th{\'e}ophile Gaudin, Peter Bolgar, Christopher~A Hunter, Costas Bekas, and Alpha~A Lee.
\newblock Molecular transformer: a model for uncertainty-calibrated chemical reaction prediction.
\newblock \emph{ACS central science}, 5\penalty0 (9):\penalty0 1572--1583, 2019.

\bibitem[Shi et~al.(2020)Shi, Xu, Zhu, Zhang, Zhang, and Tang]{shi2020graphaf}
Chence Shi, Minkai Xu, Zhaocheng Zhu, Weinan Zhang, Ming Zhang, and Jian Tang.
\newblock Graphaf: a flow-based autoregressive model for molecular graph generation.
\newblock In \emph{International Conference on Learning Representations}, 2020.
\newblock URL \url{https://openreview.net/forum?id=S1esMkHYPr}.

\bibitem[Simonovsky and Komodakis(2018)]{simonovsky2018graphvae}
Martin Simonovsky and Nikos Komodakis.
\newblock Graphvae: Towards generation of small graphs using variational autoencoders.
\newblock In \emph{Artificial Neural Networks and Machine Learning--ICANN 2018: 27th International Conference on Artificial Neural Networks, Rhodes, Greece, October 4-7, 2018, Proceedings, Part I 27}, pages 412--422. Springer, 2018.

\bibitem[Song et~al.(2021)Song, Sohl-Dickstein, Kingma, Kumar, Ermon, and Poole]{song2021scorebased}
Yang Song, Jascha Sohl-Dickstein, Diederik~P Kingma, Abhishek Kumar, Stefano Ermon, and Ben Poole.
\newblock Score-based generative modeling through stochastic differential equations.
\newblock In \emph{International Conference on Learning Representations}, 2021.
\newblock URL \url{https://openreview.net/forum?id=PxTIG12RRHS}.

\bibitem[Sutton and Barto(2018)]{sutton2018reinforcement}
Richard~S Sutton and Andrew~G Barto.
\newblock \emph{Reinforcement learning: An introduction}.
\newblock MIT press, 2018.

\bibitem[Tolstikhin et~al.(2018)Tolstikhin, Bousquet, Gelly, and Schoelkopf]{tolstikhin2018wasserstein}
Ilya Tolstikhin, Olivier Bousquet, Sylvain Gelly, and Bernhard Schoelkopf.
\newblock Wasserstein auto-encoders.
\newblock In \emph{International Conference on Learning Representations}, 2018.
\newblock URL \url{https://openreview.net/forum?id=HkL7n1-0b}.

\bibitem[Vaswani et~al.(2017)Vaswani, Shazeer, Parmar, Uszkoreit, Jones, Gomez, Kaiser, and Polosukhin]{vaswani2017attention}
Ashish Vaswani, Noam Shazeer, Niki Parmar, Jakob Uszkoreit, Llion Jones, Aidan~N Gomez, {\L}ukasz Kaiser, and Illia Polosukhin.
\newblock Attention is all you need.
\newblock \emph{Advances in neural information processing systems}, 30, 2017.

\bibitem[Vignac et~al.(2023)Vignac, Krawczuk, Siraudin, Wang, Cevher, and Frossard]{vignac2023digress}
Clement Vignac, Igor Krawczuk, Antoine Siraudin, Bohan Wang, Volkan Cevher, and Pascal Frossard.
\newblock Digress: Discrete denoising diffusion for graph generation.
\newblock In \emph{The Eleventh International Conference on Learning Representations}, 2023.
\newblock URL \url{https://openreview.net/forum?id=UaAD-Nu86WX}.

\bibitem[Weininger(1988)]{weininger1988smiles}
David Weininger.
\newblock Smiles, a chemical language and information system. 1. introduction to methodology and encoding rules.
\newblock \emph{Journal of chemical information and computer sciences}, 28\penalty0 (1):\penalty0 31--36, 1988.

\bibitem[Williams(1992)]{williams1992simple}
Ronald~J Williams.
\newblock Simple statistical gradient-following algorithms for connectionist reinforcement learning.
\newblock \emph{Machine learning}, 8:\penalty0 229--256, 1992.

\bibitem[Winter et~al.(2019)Winter, Montanari, Steffen, Briem, No{\'e}, and Clevert]{winter2019efficient}
Robin Winter, Floriane Montanari, Andreas Steffen, Hans Briem, Frank No{\'e}, and Djork-Arn{\'e} Clevert.
\newblock Efficient multi-objective molecular optimization in a continuous latent space.
\newblock \emph{Chemical science}, 10\penalty0 (34):\penalty0 8016--8024, 2019.

\bibitem[Xie et~al.(2021)Xie, Shi, Zhou, Yang, Zhang, Yu, and Li]{xie2021mars}
Yutong Xie, Chence Shi, Hao Zhou, Yuwei Yang, Weinan Zhang, Yong Yu, and Lei Li.
\newblock {\{}MARS{\}}: Markov molecular sampling for multi-objective drug discovery.
\newblock In \emph{International Conference on Learning Representations}, 2021.
\newblock URL \url{https://openreview.net/forum?id=kHSu4ebxFXY}.

\bibitem[Yan et~al.(2017)Yan, Zhang, Zhou, Li, and Huang]{yan2017hdock}
Yumeng Yan, Di~Zhang, Pei Zhou, Botong Li, and Sheng-You Huang.
\newblock Hdock: a web server for protein--protein and protein--dna/rna docking based on a hybrid strategy.
\newblock \emph{Nucleic acids research}, 45\penalty0 (W1):\penalty0 W365--W373, 2017.

\bibitem[Yang et~al.(2022)Yang, Zeng, Wu, Jia, and Yan]{yang2022learning}
Nianzu Yang, Kaipeng Zeng, Qitian Wu, Xiaosong Jia, and Junchi Yan.
\newblock Learning substructure invariance for out-of-distribution molecular representations.
\newblock \emph{Advances in Neural Information Processing Systems}, 35:\penalty0 12964--12978, 2022.

\bibitem[Yang et~al.(2023)Yang, Zeng, Wu, and Yan]{yang2023molerec}
Nianzu Yang, Kaipeng Zeng, Qitian Wu, and Junchi Yan.
\newblock Molerec: Combinatorial drug recommendation with substructure-aware molecular representation learning.
\newblock In \emph{Proceedings of the ACM Web Conference 2023}, pages 4075--4085, 2023.

\bibitem[Yang et~al.(2021{\natexlab{a}})Yang, Hwang, Lee, Ryu, and Hwang]{yang2021hit}
Soojung Yang, Doyeong Hwang, Seul Lee, Seongok Ryu, and Sung~Ju Hwang.
\newblock Hit and lead discovery with explorative rl and fragment-based molecule generation.
\newblock \emph{Advances in Neural Information Processing Systems}, 34:\penalty0 7924--7936, 2021{\natexlab{a}}.

\bibitem[Yang et~al.(2021{\natexlab{b}})Yang, Ouyang, Dang, Zheng, Li, and Zhou]{yang2021knowledge}
Yuwei Yang, Siqi Ouyang, Meihua Dang, Mingyue Zheng, Lei Li, and Hao Zhou.
\newblock Knowledge guided geometric editing for unsupervised drug design.
\newblock \emph{openreview.net}, 2021{\natexlab{b}}.

\bibitem[You et~al.(2018)You, Liu, Ying, Pande, and Leskovec]{you2018graph}
Jiaxuan You, Bowen Liu, Zhitao Ying, Vijay Pande, and Jure Leskovec.
\newblock Graph convolutional policy network for goal-directed molecular graph generation.
\newblock \emph{Advances in neural information processing systems}, 31, 2018.

\bibitem[Zhang et~al.(2019)Zhang, Tong, Xu, and Maciejewski]{zhang2019graph}
Si~Zhang, Hanghang Tong, Jiejun Xu, and Ross Maciejewski.
\newblock Graph convolutional networks: a comprehensive review.
\newblock \emph{Computational Social Networks}, 2019.

\bibitem[Zhavoronkov et~al.(2019)Zhavoronkov, Ivanenkov, Aliper, Veselov, Aladinskiy, Aladinskaya, Terentiev, Polykovskiy, Kuznetsov, Asadulaev, et~al.]{zhavoronkov2019deep}
Alex Zhavoronkov, Yan~A Ivanenkov, Alex Aliper, Mark~S Veselov, Vladimir~A Aladinskiy, Anastasiya~V Aladinskaya, Victor~A Terentiev, Daniil~A Polykovskiy, Maksim~D Kuznetsov, Arip Asadulaev, et~al.
\newblock Deep learning enables rapid identification of potent ddr1 kinase inhibitors.
\newblock \emph{Nature biotechnology}, 2019.

\bibitem[Zhu(2020)]{zhu2020big}
Hao Zhu.
\newblock Big data and artificial intelligence modeling for drug discovery.
\newblock \emph{Annual review of pharmacology and toxicology}, 2020.

\bibitem[Zhu et~al.(2017)Zhu, Park, Isola, and Efros]{zhu2017unpaired}
Jun-Yan Zhu, Taesung Park, Phillip Isola, and Alexei~A Efros.
\newblock Unpaired image-to-image translation using cycle-consistent adversarial networks.
\newblock In \emph{Proceedings of the IEEE international conference on computer vision}, pages 2223--2232, 2017.

\end{thebibliography}


\end{document}